\documentclass[11pt,a4paper]{article}
\usepackage[hyperref]{eacl2021}
\usepackage{times}
\usepackage{latexsym}
\usepackage{multirow}
\usepackage{CJKutf8}

\usepackage{comment}
\usepackage{graphicx}
\usepackage{textcomp}
\usepackage{caption}
\usepackage{subcaption}

\usepackage{microtype}

\usepackage{booktabs}

\aclfinalcopy 
\def\aclpaperid{39} 

\title{Emoji-Based Transfer Learning for Sentiment Tasks}

\author{Susann Boy \\
  Saarland University \\\And
  Dana Ruiter \\
  Saarland University \\
  \texttt{\{sboy,druiter,dietrich.klakow\}@lsv.uni-saarland.de} \\\And
  Dietrich Klakow \\
  Saarland University \\
  }

\date{}

\begin{document}
\maketitle
\begin{abstract}
Sentiment tasks such as hate speech detection and sentiment analysis, especially when performed on languages other than English, are often low-resource. In this study, we exploit the emotional information encoded in emojis to enhance the performance on a variety of sentiment tasks. This is done using a transfer learning approach, where the parameters learned by an emoji-based source task are transferred to a sentiment target task. We analyse the efficacy of the transfer under three conditions, i.e. $i)$ the emoji content and $ii)$ label distribution of the target task as well as $iii)$ the difference between monolingually and multilingually learned source tasks.
We find i.a. that the transfer is most beneficial if the target task is balanced with high emoji content. Monolingually learned source tasks have the benefit of taking into account the culturally specific use of emojis and gain up to F1 $+0.280$ over the baseline.
\end{abstract}

\section{Introduction}
\label{s:introduction}

Many natural language processing (NLP) tasks suffer from a lack of available data. This is especially true for sentiment tasks, such as hate speech (HS) detection, which depend on the availability of manually annotated data. 
When moving to languages other than English, many sentiment tasks quickly become very low-resourced. 

On the other hand, noisy social media content is available in abundance and many sentiment tasks are based on user comments on such platforms. Emojis can be a valuable source for the distant supervision of sentiment tasks, as they correlate with the underlying emotion of a comment. In this study, we aim to exploit the emotional information encoded in emojis to improve the performance on various sentiment tasks using a transfer learning approach from an emoji-based \textbf{source task} (ST) to a sentiment \textbf{target task} (TT). Previous work has focused on the transfer from predicting single emojis \citep{felbo-etal-2017-using} or strictly pre-defined emoji-clusters \citep{deriu-etal-2016-swisscheese}. However, pre-defined emoji clusters do not take into account the culturally diverse usage of emojis \citep{park2012crosscultural,kaneko2019emojigrid}. We therefore introduce data-driven supervised and unsupervised emoji clusters and compare these with single emoji prediction tasks. Specifically, we analyze the efficacy of the transfer from a single emoji or (un)supervised emoji cluster prediction ST to a sentiment TT under three conditions, i.e. \textit{i)} low vs. high amount of \textbf{emoji content} present in TT, \textit{ii)} balanced vs. unbalanced \textbf{label distribution} in TT and \textit{iii)} \textbf{monolingually} or \textbf{multilingually} learned ST. The first two conditions are based on typical qualities of sentiment corpora, which tend to be unbalanced in their label distribution with varying degrees of emoji content depending on the source of the data. The third condition is relevant for languages for which a TT is low-resource and which might benefit from a multilingually learned ST.

In Section \ref{s:related_work} we give an outline of related work, followed by the introduction of our method (Section \ref{s:method}). The experimental setup in Section \ref{s:experimental_setup} details the data and models used as well as the (un)supervised clusters generated. In Section \ref{s:results} we describe our results and conclude in Section \ref{s:discussion}.

\section{Related Work}
\label{s:related_work}

\textbf{Emojis} have been used as a type of distant supervision using pre-defined emotion classes based on psychological models \citep{suttles2013distant}, binary (\textit{positive}/\textit{negative}) classes \citep{deriu-etal-2016-swisscheese} or a set of single emojis \citep{felbo-etal-2017-using}. However, such pre-defined emoji classes often do not account for the culturally diverse use of emojis \citep{park2012crosscultural,kaneko2019emojigrid}. In contrast, our work does not pre-define the emotion classes found in emojis and instead learns these classes, or clusters, from the data itself. While our and the above approaches focus on exploiting emojis as additional labelled data, e.g. in a transfer setting, emoji embeddings \citep{eisner-etal-2016-emoji2vec} have been used as additional features in downstream tasks such as sarcasm detection \citep{subramanian2019exploiting}.

\textbf{Transfer learning} has recently been driven by transformer-based \citep{vaswani2017attention} language models (LM) such as BERT \citep{devlin-etal-2019-bert} or XLM-R \citep{conneau-etal-2020-unsupervised}. When learning a source task on these models, the representations in the encoder change to become informative to the task at hand. In a parameter transfer setting, a new but related target task then profits from the learned representations in the encoder. Transfer learning has been applied to sentiment analysis (SA) using parameter transfer methods such as pre-trained sentiment embeddings \citep{dong-de-melo-2018-helping} or machine translation-based context vectors \citep{mccann2017learned}. Our approach forms part of the parameter transfer approach, as we use encoder representations learned using emoji-based source tasks and transfer these to sentiment target tasks.

\textbf{Hate speech} classification and \textbf{sentiment analysis} have in recent years been the object of many shared tasks \citep{rosenthal2017semeval,ruppendorfer2018germeval,basile2019semeval,hasoc2019,ogr:kob:2019:poleval}. Classification models for these tasks often rely on feature engineering and statistical methods such as naive-bayes \cite{saleem_web_2016}, logistic regression over subwords \cite{waseem2016hateful} or neural approaches including convolutional neural networks \cite{park2017one} or, as in our case, the representations of large LMs \citep{yang2019xlnet}. 

\section{Method: Emoji-Prediction}
\label{s:method}

For our parameter transfer, we rely on a single transformer-based LM which is shared among different tasks.
A sequence $x \in X$ is featurized by reading it into the encoder of the LM and retrieving its last hidden state. A linear layer is then used as a predictive function $f:X \rightarrow Y$ to predict labels $y \in Y$. A task $\mathcal{T}=\{Y, f(x)\}$ is then a set of labels $Y$ and the predictive function $f$ over the instances in $X$. 

We follow a \textbf{transfer learning} approach, where source task $\mathcal{T}_S$ is an emoji-based classification task, i.e. given a sequence, predict the emoji (class) that it originally contained. Target task $\mathcal{T}_T$ is a downstream task such as SA or HS (Section \ref{s:data}). Each task has its own set of instances $X$, labels $Y$ and predictive function $f$, while the feature-generating LM stays the same. The error of predictor $f$ is back-propagated to the LM, which allows us to transfer learned parameters from $\mathcal{T}_S$ to $\mathcal{T}_T$.

\subsection{Source Tasks (ST)} We focus on 5 different emoji-based STs, that can be divided into two types, emoji prediction (EP) and emoji cluster prediction. To sample emojis for EP or create clusters, we rely on a large collection of user generated comments.
\textbf{EP} is a multi-class prediction task over the 64 most common emojis identified in the collection of comments. Concretely, given a tweet with all emojis removed, the classifier has to predict which of the 64 emojis was originally contained within it.

The \textbf{emoji cluster prediction} tasks can be supervised (PMI-\{Target,Swear\}) or unsupervised (KMeans-\{2,3\}). In this case the task is simplified: Given a tweet with all emojis removed, predict the cluster to which the emoji originally contained in the tweet belonged.

\paragraph{Unsupervised Clusters} In order to account for the cultural differences in the use of emojis, we learn emoji clusters directly from the user generated data. We generate 50-dimensional vector representations over the tokens in the collection of user comments using the continuous bag of words \citep{mikolov2013efficient} approach. We then perform k-means clustering with 6 target clusters on the representations of emojis that occurred $\geq 1000$ times. These clusters are manually merged into 2 (\textit{positive}/\textit{negative}) and 3 (\textit{positive}/\textit{negative}/\textit{neutral}) clusters to create the binary \textbf{KMeans-2} and ternary \textbf{KMeans-3} emoji cluster prediction STs respectively.
Below a comment to be classified as \emph{positive} according to the KMeans-\{2,3\} tasks, as it originally contained an emoji that belonged to the \emph{positive} cluster:

\begin{quote}
\textit{So beautiful and great advice} \textrightarrow \textit{positive}
\end{quote}

\paragraph{Supervised Clusters} As an alternative to the completely unsupervised clusters, we exploit the mutual information between emojis and swear words as a type of distant supervision for HS tasks. We calculate the pointwise mutual information (PMI) between comments in our collection of user content (not) containing slurs and the emojis that appear. An emoji is in the slur cluster if its PMI is larger to comments containing swearwords, otherwise it is in the neutral cluster. \textbf{PMI-Swear} is then a binary classification task based on the resulting slur/neutral emoji clusters.

While the unsupervised emoji cluster prediction STs and PMI-Swear are source-oriented, i.e. learned on user generated content, we also explore target-oriented clusters that rely on the shared information between emojis and the labels in each of the TTs.
Concretely, we calculate the PMI between the label of an instance in the respective TT training data and the emojis it contains. The emoji is placed into the cluster of the label to which its PMI value is largest. \textbf{PMI-Target} is the ST based on these target-oriented emoji clusters.

\subsection{Target Tasks (TT)} Once the classifier has been fully trained on the ST, and thus has adapted the underlying LMs representations to fit the ST at hand, we discard it and train a new classifier on top of the enriched LM to predict the TT. We evaluate this transfer from the various STs on two main categories of TTs, namely Hate Speech Detection and Sentiment Analysis. Given a user generated comment, \textbf{Hate Speech} Detection is the task of classifying the comment as either \textit{hate} or \textit{none}. Note, however, that concrete label names (e.g. \textit{offense}, \textit{hate}, \textit{harmful}) may differ across specific HS tasks. 

While HS in our case is a binary classification task, \textbf{Sentiment Analysis} is a ternary classification task which takes as input a user generated comment and classifies it as either \emph{positive}, \emph{neutral} or \emph{negative}. In the following an example from the Sentiment Analysis in Twitter \citep{rosenthal2017semeval} task:

\begin{quote}
    \textit{Finally starting the 5th season of \#Dexter. See ya later, weekend!} \textrightarrow \textit{positive}
\end{quote}

Both HS and SA are sentiment-based tasks, e.g. \emph{hate} towards a group of people or \emph{positive} sentiment towards a product etc. We therefore take these two types of tasks to have the potential to benefit from the emotion information encoded in emojis. In the following sections we explore the conditions under which the transfer from an emoji-based ST to a sentiment-based TT is beneficial for the TT.

\section{Experimental Setup}
\label{s:experimental_setup}

We describe the data used for the STs and TTs respectively (Section \ref{s:data}), followed by the specifications of the encoding LM (Section \ref{s:model_specs}) and the emoji cluster creation (Section \ref{s:clusters}).

\subsection{Data}
\label{s:data}

\begin{table}[t]
\small
\centering
\setlength\tabcolsep{2pt}
\begin{tabular}{l  r r r}
\toprule
Corpus      & \multicolumn{2}{c}{\# Tweets}  &  \# Emojis  \\ 
     & Train  & Test  &    \\ 
      \midrule
\multicolumn{2}{l}{\emph{Target Tasks (TT)}} \\
HS-DE   & 1158/2439  & 970/2061 & 853 (7.2\%)\\ 
SA-DE & 1346/900/3676  & 83/49/197 & 166 (2\%)\\ 
HS-ES & 1857/2643  & 660/940 & 957 (14.5\%)\\
SA-EN & 18481/7551/21542  & 2375/3972/5937 & 1211 (1.9\%)\\
SA-AR & 653/1022/1336  & 1514/2222/2364 & 2126 (22.5\%)\\
HS-PL & 812/8726  & 134/866 & 1733 (13.7\%) \\
\midrule
\multicolumn{2}{l}{\emph{Source Tasks (ST)}} \\
TW-DE & 16M & -- & 3M (10\%)\\
TW-EN & 323M & -- & 82M (17\%) \\
TW-ES & 320M & -- & 43M (9\%) \\
TW-PL & 7M & -- & 1M  (12\%) \\
TW-AR & 183M & -- & 56M (20\%)\\ 
\bottomrule
\end{tabular}
\caption{Number of train, test (for TT) and collected (for ST) tweets as well as number of (non-unique) emojis contained in each corpus. Percentage of training tweets containing emojis in brackets. TTs with label distribution for HS (\textit{hate}/\textit{none}) and SA (\textit{positive}/\textit{negative}/\textit{neutral}) tasks.}
\label{t:corpora}
\end{table}

\paragraph{Source Tasks}

We use a collection\footnote{\url{www.archive.org/details/twitterstream}} of tweets that has been collected from the Twitter stream between 2011 and 2019 as our corpus needed to sample emojis and create emoji clusters for the STs. We perform language identification using the \texttt{polyglot}\footnote{\url{www.github.com/aboSamoor/polyglot}} library
over the tweets to create a corpus for German, English, Spanish, Polish and Arabic (TW-\{DE,EN,ES,PL,AR\}) respectively.

To automatically identify swear words for PMI-Swear, we use a German and a multilingual swear word collection, namely \texttt{WoltLab}\footnote{\url{www.woltlab.com/attachment/3615-schimpfwortliste-txt/}} and \texttt{Hatebase}\footnote{\url{www.hatebase.org/}}. In total, we collected 785 slurs for German, and 1531, 140, 306, 79 for English, Spanish, Polish and Arabic respectively.

\paragraph{Target Tasks}

We work with 6 target tasks in total, 3 HS and 3 SA tasks, taking into account their emoji content, class (im)balance and language.

For German, we use GermEval 2018 \citep{ruppendorfer2018germeval} Task 1 (\textit{offense}/\textit{other}) (HS-DE) and SB10k \citep{cieliebak-etal-2017-twitter} (\textit{positive}/\textit{negative}/\textit{neutral}) (SA-DE). For English, we use Sentiment Analysis in Twitter \citep{rosenthal2017semeval} (\textit{positive}/\textit{negative}/\textit{neutral}) (SA-EN). Sentiment Analysis in Twitter is also used for Arabic (SA-AR). For Spanish we use HatEval \citep{basile2019semeval} (\textit{hate}/\textit{none}) (HS-ES) and for Polish, we use PolEval \citep{ogr:kob:2019:poleval} Task 6 (\textit{harmful}/\textit{none}) (HS-PL). For all of the above, we use the original train/test splits. While the HA tasks have different label names, we normalize these to be \textit{hate}/\textit{none} across all tasks. For all SA, the labels to be predicted are \textit{positive}/\textit{negative}/\textit{neutral}.

In Table \ref{t:corpora}, we report the label distribution, \textit{hate}/\textit{none} for HS and \textit{positive}/\textit{negative}/\textit{neutral} for SA, across all TT training and test sets, as well as ST Twitter corpora sizes. For both ST and TT corpora, we also report the percentage as well as total number of tweets containing emojis.

\paragraph{Preprocessing}

All data sets undergo the same preprocessing. Tweets are tokenized using the NLTK \citep{loper02nltk} \texttt{TweetTokenizer} and user mentions, 
retweets and punctuation are removed. Repeated characters are shortened. We use token frequencies to determine the standard orthography of a word (e.g. \emph{coooool} \textrightarrow\  \emph{cool} instead of \emph{col}).

\subsection{Model Specifications}
\label{s:model_specs}

For the monolingual (German) experiments, we use the German BERT\footnote{\url{www.deepset.ai/german-bert}} (BERT-DE) and for multilingual experiments we use \texttt{Bert-Base-Multilingual-Cased} (BERT-M) as the LM to encode the tweets.
We base our code\footnote{\url{https://github.com/uds-lsv/emoji-transfer}} on the \texttt{simpletransformers}\footnote{\url{www.github.com/ThilinaRajapakse/simpletransformers}} sequence classification implementations of the above models. Each classification task is trained for a maximum of 10 epochs using early stopping over the validation accuracy with $\delta=0.01$ and patience 3. Training was performed on a single Titan-X GPU, which took between 1 and 6 hours depending on the data size. We evaluate the resulting classifiers using the Macro F1 measure.

\subsection{Clusters}
\label{s:clusters}

We describe the creation of the emoji clusters used for the emoji cluster STs.

\begin{figure}[t]
    \centering
    \includegraphics[width=\columnwidth]{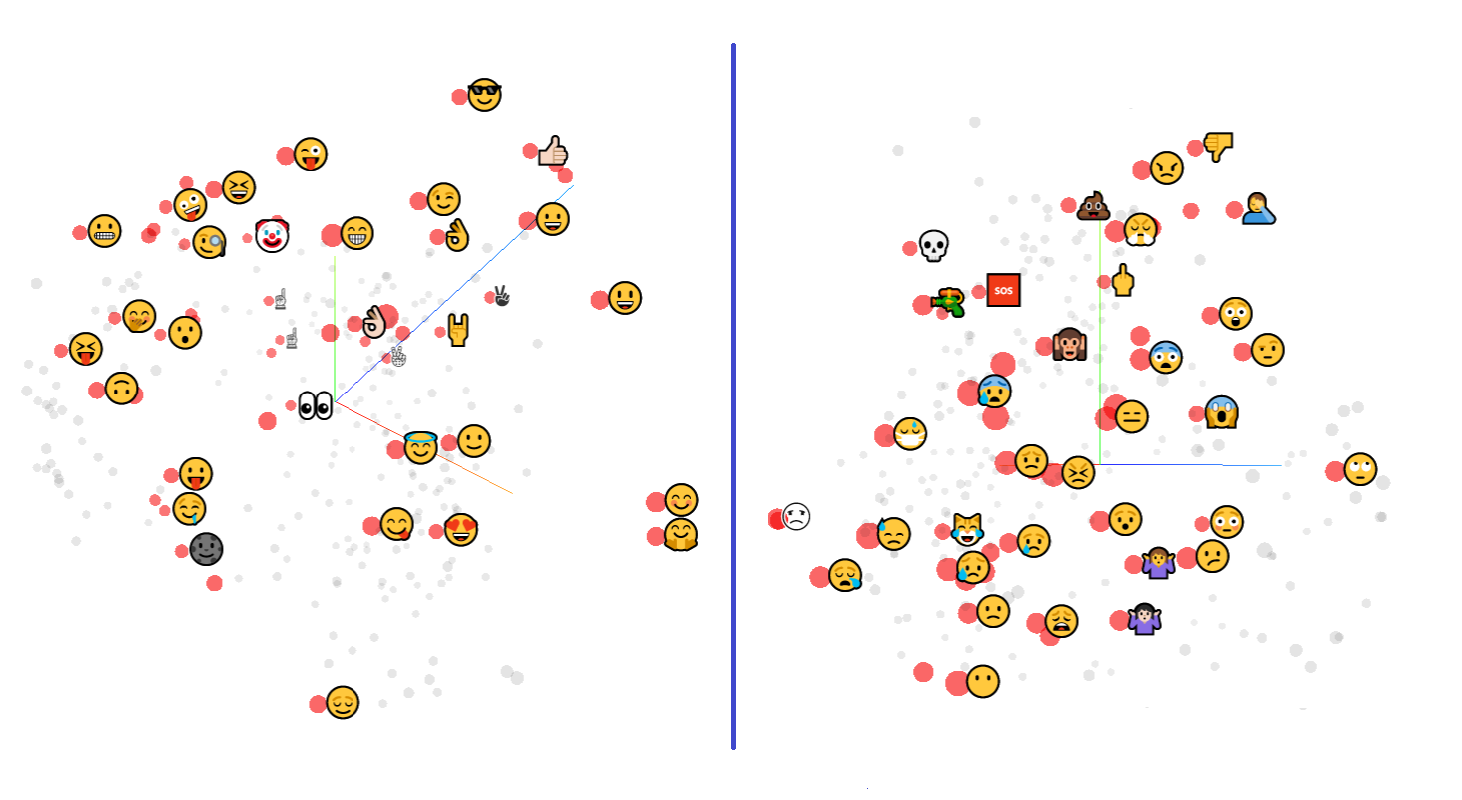}
    \caption{\emph{Happy} (left) and \emph{unhappy} (right) emoji clusters obtained by KMeans on TW-DE.}
    \label{f:w2v_cluster}
\end{figure}

\paragraph{Unsupervised} The unsupervised clusters (Section \ref{s:method}) were trained on TW-DE and the concatenation of TW-\{DE,EN,ES,PL,AR\} for the mono- and multilingual experiments respectively.
In both cases, this yielded clusters that can be manually categorized as \emph{happy, love, fun, nature, unhappy, other} (Figure \ref{f:w2v_cluster}). For KMeans-3, \{\emph{happy, fun, love}\} were merged to \emph{positive}, \{\emph{other, nature}\} to \emph{neutral} and \{\emph{unhappy}\} was used as the \emph{negative} class. For KMeans-2, the \emph{neutral} class is ignored.

\paragraph{Supervised} The PMI-Target clusters are trained on the respective TT training data. 
The slur lists are used to identify the slurs in the twitter corpora. PMI-Swear is then trained on TW-DE and the concatenation of TW-\{DE,EN,ES,PL,AR\} for the mono- and multilingual experiments respectively.

\section{Results}
\label{s:results}

\begin{figure*}[t]
    \centering
    \includegraphics[width=\textwidth]{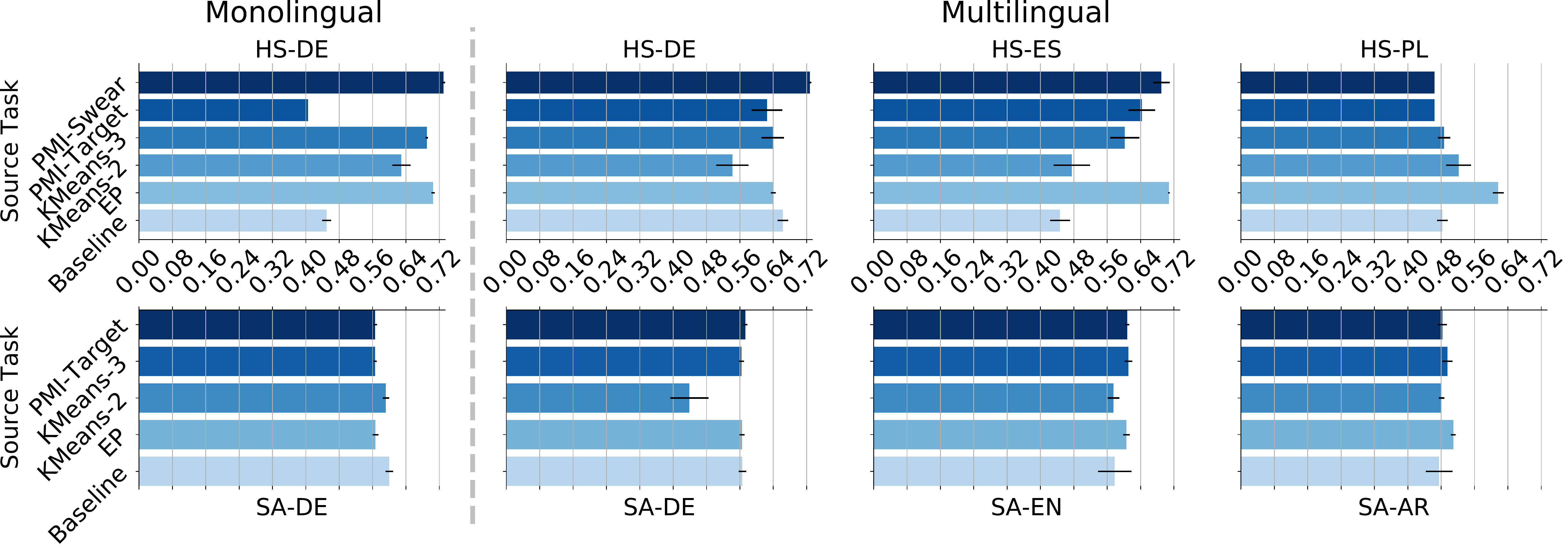}
    \caption{Macro F1 of the HS and SA target tasks transferred from monolingual (left) and multilingual (right) STs.
    }
    \label{f:results}
\end{figure*}

We train each model over 10 seeded runs and report the averaged Macro F1 with standard error (Figure \ref{f:results}). For each TT, we train a \textbf{baseline}, which is the same pre-trained BERT-\{DE,M\} model that is now fine-tuned directly on the TT classification task at hand, without prior training on the ST. We compare these baselines with those models that have undergone a transfer from ST to TT.
We use the term \emph{equivalent} to signify that two models lie within each others error bounds.

\subsection{Condition 1: Emoji Content}
\label{s:emoji_content}

We evaluate the effect that STs have on TTs with different amounts of emoji content. We focus on the TTs with the lowest and highest amount of emoji content, namely SA-EN (1.9\% emoji content) and SA-AR (22.5\%). This is the multilingual case.
For the monolingual case, we evaluate the effect on SA-DE (2\%) and HS-DE (7.2\%). All of these TTs are unbalanced, i.e. the minority class makes up 15.2--32.2\% of the training data.

The \textbf{monolingual}, low emoji content SA-DE task does not profit from the transfer. Rather, the training on most STs leads to a slight drop in F1-Macro compared to the baseline (F1 0.600). On the other hand, high emoji content HS-DE greatly benefits from the transfer, with PMI-Swear (F1 0.730) being especially beneficial for the performance on the TT, yielding a gain of F1 +0.280 over the baseline. 
This shows that the shared information in emojis and slurs is relevant to the HS task at hand. 
Also beneficial are EP (F1 0.705), and the unsupervised KMeans-3 (F1 0.690) and KMeans-2 (F1 0.629) cluster prediction tasks. Only the supervised PMI-Target (F1 0.405) does no seem to be beneficial for the performance on the TT, leading to a drop in performance, which is due to the unbalanced nature of the TT (Section \ref{s:label_dist}).

The \textbf{multilingual} case shows a slightly mixed trend. Low emoji content SA-EN does not benefit from the transfer, but unlike in the monolingual setting, it is not harmed by it either. All STs lead to a TT performance that is equivalent to the baseline (F1 0.578). 
High emoji content SA-AR only barely profits from the transfer, with EP (F1 0.509) leading to a small gain of F1 (+0.034) over the baseline (F1 0.475), while all other STs lead to an equivalent performance to the baseline. The overall trend is similar to the monolingual case but the positive and negative effects are dimmed down, which may be due to the multilingual aspect (Section \ref{s:multilinguality}).

The \textbf{general trend} shows that a decent amount of emoji content in the TT training data is crucial for the transfer to be beneficial.

\subsection{Condition 2: Label Distribution}
\label{s:label_dist}

To analyze the effect that the STs have on differently (un)balanced TTs, we focus on HS-PL (the minority class makes up 8.5\% of training data) and HS-ES (41.3\%), as they are the two most (un)balanced TTs, while being comparable in terms of emoji content (13.7\% and 14.5\% respectively).

For \textbf{unbalanced} HS-PL, EP (F1 0.617) and unsupervised KMeans-2 (F1 0.522) lead to an improvement of F1 $+0.134$ and F1 $+0.039$ over the baseline, respectively. All other STs are equivalent to the baseline. \textbf{Balanced} HS-ES benefits from all TTs, with EP (F1 0.708) leading to a gain of F1 $+0.261$ over the baseline (F1 0.447), followed by PMI-Swear (F1 0.690) and PMI-Target (F1 0.643). The unsupervised clusters are beneficial but less effective, with F1 0.602 and F1 0.475 for KMeans-3 and KMeans-2 respectively, which likely stems from the multilingual aspect (Section \ref{s:multilinguality}). 

\textbf{PMI-Target} performs poorly on unbalanced HS-PL (and HS-DE etc.) due to its use of mutual information between emojis and the TT labels. This leads to it reproducing the class imbalance, making it less effective on unbalanced TTs.

The difference in impact of \textbf{PMI-Swear} on HS-PL (none) and HS-ES  (and HS-DE) (gain) can be explained by the composition of the ST dataset. TW-PL is the smallest corpus in the multilingual collection of user comments, and this sparsity is further driven by the morphological complexity of Polish, such that the 306 slurs from the Polish slur list only resulted in 65$k$ Polish training samples in PMI-Swear, as opposed to 1.8M and 3M for German and Spanish respectively.

\textbf{Overall}, if the label distribution in TT is balanced, the TT easily benefits from the transfer. Otherwise other conditions such as the multilinguality or emoji content become more relevant.

\subsection{Condition 3: Multilinguality}
\label{s:multilinguality}

We analyze the effectiveness of the transfer in a monolingual and multilingual setting. For this, we focus on the effect that the monolingually and multilingually learned STs have on HS-DE and SA-DE. Both TTs are unbalanced, while HS-DE has a high emoji content and SA-DE has a low emoji content.

The different effects of the emoji-content in HS-DE and SA-DE has been discussed in Section \ref{s:emoji_content}, showing that in the \textbf{monolingual} setting, high emoji content HS-DE benefits from the transfer, while low emoji content SA-DE does not. In the \textbf{multilingual} case, we see a similar, but dimmed, trend. SA-DE does not benefit from the transfer, with all TTs leading to an equivalent performance as the baseline (F1 0.566), except KMeans-2 (F1 0.439) which is below the baseline. The STs have a similar performance on HS-DE, being equivalent or below the baseline (F1 0.663). Only PMI-Swear (F1 0.678) is beneficial for the TT performance.

The effect of ST-oriented clusters KMeans-\{2,3\} was beneficial in the monolingual case (HS-DE), but this benefit is lost in the multilingual setting.
This underlines our original idea that ST-oriented unsupervised emoji clusters learned on large amounts of user generated text have the advantage of accounting for \textbf{cultural differences} in the usage of emojis.
When learned multilingually, this advantage is lost.
An example of the culturally diverse use of emojis is \includegraphics[scale=0.27]{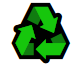}, which is rather infrequent in Europe and might be used to point towards the importance of \emph{recycling}. In TW-AR, this emoji is among the top 5 most frequent emojis, and is used to
motivate other users to \emph{share} their content. 

The \textbf{overall trend} thus shows that monolingually learned STs are more beneficial than multilingual STs. However, if the training data of a TT is balanced, this effect is less pronounced.

\subsection{Comparison to Benchmark Results}

\begin{table}[t]
\small
\centering
\setlength\tabcolsep{4pt}
\begin{tabular}{l  l r r }
\toprule
TT     & Method  & F1  &    SOTA  \\ 
      \midrule
HS-DE & PMI-Swear (monolingual) & 0.730  & \textbf{0.768} \\
HS-ES & EP & 0.708 & \textbf{0.730} \\
HS-PL & EP & \textbf{0.617} & 0.586 \\
\midrule
SA-DE & Baseline (monolingual) & 0.600  &\textbf{ 0.651} \\
SA-AR & EP & 0.509 & \textbf{0.610} \\
SA-EN & KMeans-3 & 0.611 & \textbf{0.677} \\
\bottomrule
\end{tabular}
\caption{Macro F1 comparison of top-scoring transfer method (\textit{F1}) with SOTA results on the different TT test sets. 
Best scores in \textbf{bold}. 
See \citep{montani2018tuwienkbs} (HS-DE), \citep{basile2019semeval} (HS-ES), \citep{ogr:kob:2019:poleval} (HS-PL), \citep{cieliebak-etal-2017-twitter} (SA-DE) and \citep{rosenthal2017semeval} (HS-\{AR,EN\}) for SOTA method descriptions.
}
\label{t:benchmarks}
\end{table}

To put the results into a broader perspective, we compare to state-of-the-art (SOTA) models for each of the shared-tasks/datasets that our TTs are based on (Table \ref{t:benchmarks}).
For two of the \textbf{Hate Speech} benchmarks, the performance of our transfer approach is close to the SOTA, namely with a difference of F1 $-0.038$ (HS-DE) and F1 $-0.03$ (HS-ES). For HS-PL, we were able to achieve a gain of $+0.031$ over the SOTA.
Across all three \textbf{Sentiment Analysis} benchmarks, our models are below the SOTA.
This indicates that SA, in general, is a more difficult task to our transfer approach than HS, possibly due to its ternary, rather than binary, classification objective. This is another factor causing the transfer to be overall more beneficial for HS rather than SA, next to the unbalanced (SA-\{EN,AR\}) and low-emoji content (SA-DE) nature of the SA tasks.

\section{Summary}
\label{s:discussion}

We have evaluated and identified conditions under which the transfer from an emoji-based ST is beneficial for a sentiment TT. In the experiments in Section \ref{s:results} we observed three major trends, namely $i)$ TTs with high amounts of emoji content benefit more from the transfer, $ii)$ PMI-Target tends to be detrimental to unbalanced TTs and $iii)$ monolingually learned STs tend to perform better than their multilingual counterparts, due to their improved representation of culturally unique emoji usages. The latter underlines the importance of taking into account cultural differences when exploiting the information encoded in emojis.

From these results, we can draw conclusions about the conditions under which a given emoji-based ST is beneficial. Due to the shared information between emojis and slurs, \textbf{PMI-Swear} is beneficial to HS tasks when the data that can be generated from the swear word list is decently large. \textbf{PMI-Target} is beneficial when the TT is balanced, otherwise it replicates the already existing class imbalance. Unsupervised \textbf{KMeans-\{2,3\}} should be learned monolingually to be beneficial and \textbf{EP} is a safe choice for TTs with high emoji content.

\section*{Acknowledgments}

We want to thank the anonymous reviewers as well as Thomas Kleinbauer for their valuable feedback.
The project on which this paper is based is funded by the DFG under funding code WI 4204/3-1.

\bibliography{anthology,eacl2021}
\bibliographystyle{acl_natbib}

\appendix
\end{document}